\documentclass{article}

\usepackage{iclr2027_conference,times}

\usepackage{amsmath,amsfonts,bm}

\def\eqref#1{equation~\ref{#1}}

\def\1{\bm{1}}

\DeclareMathAlphabet{\mathsfit}{\encodingdefault}{\sfdefault}{m}{sl}
\SetMathAlphabet{\mathsfit}{bold}{\encodingdefault}{\sfdefault}{bx}{n}

\newcommand{\KL}{D_{\mathrm{KL}}}

\usepackage{amsmath,amssymb,amsfonts,amsthm}
\usepackage{mathtools}

\usepackage{graphicx}
\usepackage[table]{xcolor}
\usepackage{booktabs}
\usepackage{multirow}
\usepackage{array}
\usepackage{wrapfig}
\usepackage{subcaption}
\usepackage{mdframed}
\usepackage{algorithm}
\usepackage{algorithmic}
\usepackage{float}

\definecolor{resultblue}{HTML}{E6F1FF}

\usepackage{enumitem}
\usepackage{microtype}

\usepackage{lipsum}
\usepackage{placeins}
\theoremstyle{plain}

\newmdtheoremenv[
  linewidth=0.6pt,
  linecolor=black!55,
  backgroundcolor=white,
  roundcorner=0pt,
  skipabove=7pt,
  skipbelow=7pt,
  innertopmargin=7pt,
  innerbottommargin=7pt,
  innerleftmargin=9pt,
  innerrightmargin=9pt,
  splittopskip=7pt,
  splitbottomskip=7pt
]{proposition}{Proposition}[section]

\usepackage{hyperref}
\usepackage{url}
\usepackage[capitalize]{cleveref}

\usepackage{tabularx}

\definecolor{paperblue}{HTML}{0000EE}

\hypersetup{
    colorlinks=true,
    linkcolor=paperblue,
    citecolor=paperblue,
    urlcolor=black
}
\usepackage{etoolbox}

\crefname{section}{Sec.}{Secs.}
\Crefname{section}{Section}{Sections}
\crefname{table}{Tab.}{Tabs.}
\Crefname{table}{Table}{Tables}

\usepackage{xspace}
\newcommand{\method}{CGPI\xspace}

\newcommand{\stopact}{\mathrm{STOP}}

\providecommand{\sg}{\mathrm{sg}}

\setcitestyle{numbers,square,comma}



\title{
Credit-Guided Policy Improvement for Test-time Adaptive Vision-Language Navigation
}

\author{
Yang Li\textsuperscript{1} \quad
Sijia Zhang\textsuperscript{1} \quad
Yihan Li\textsuperscript{3} \quad
Aming Wu\textsuperscript{2} \quad
Zihao Zhang\textsuperscript{1} \quad
Ziju Han\textsuperscript{1} \quad
Yahong Han\textsuperscript{1}\thanks{Corresponding author.}
\\[0.6em]
\textsuperscript{1}School of Artificial Intelligence, Tianjin University, China
\\
\textsuperscript{2}School of Computer Science and Information Engineering, Hefei University of Technology, China
\\
\textsuperscript{3}The University of Hong Kong, Hong Kong SAR, China
\\[0.4em]
{\tt\small
\{liyang1389, 3024244296, zhangzihao2490, hanziju, yahong\}@tju.edu.cn
}
\\
{\tt\small amwu@hfut.edu.cn \quad u3678938@connect.hku.hk}
}

\iclrfinalcopy

\makeatletter
\patchcmd{\@maketitle}
  {\lhead{Published as a conference paper at ICLR 2027}}
  {\lhead{Preprint. }}
  {}{}
\makeatother

\begin{document}

\maketitle


\begin{abstract}

To advance the development of embodied general navigation, Test-time Adaptation for Vision-Language Navigation (TTA-VLN) has attracted increasing attention, aiming to adapt pretrained policies online to previously unseen environments using only test-time observations and interaction history.
However, distribution shifts in unseen environments can distort the pretrained policy's local
action preferences and lead to off-course decisions.
Existing methods seek to correct such deviations using test-time signals,
such as predictive uncertainty, trajectory-level feedback, or accumulated
adaptation experience.
Yet these signals do not directly establish whether an executed behavior
actually contributes to instruction-guided progress toward the goal.
Moreover, even when a test-time signal suggests a plausible corrective
direction, the resulting policy change may still be unreliable and should
not necessarily persist in subsequent decisions.
The central challenge is therefore twofold: how to identify whether an
interaction supports goal-directed improvement, and how to determine
whether the resulting policy update is worth retaining.
We observe that every executed action induces an immediate observation
transition that exposes evidence of its local consequence.
Based on this observation, we propose Credit-Guided Policy
Improvement (CGPI), which uses action-induced observation transitions
to recover signed, reference-relative decision credit without external
outcome feedback.
The recovered credit proposes a lightweight policy update, which is
verified against prior credit-supported interactions and retained only
when supported; otherwise, it is rolled back, while the pretrained
navigation policy remains frozen.
CGPI achieves consistent gains across the evaluated VLN benchmarks and
navigation backbones, while qualitative robot trials further illustrate
zero-shot sim-to-real feasibility.

\end{abstract}

\section{Introduction}

\noindent
As an important capability of embodied robots, Vision-Language Navigation (VLN) enables an agent to follow natural language instructions and navigate to target locations based on visual observations
\citep{anderson2018vision,fried2018speaker,hao2020towards,hong2021vlnbert}.
Despite substantial progress on established VLN benchmarks, pretrained
navigation policies can still suffer significant performance degradation when
deployed with previously unseen environments and instructions.
Test-Time Adaptation for VLN (TTA-VLN)
\citep{gao2024fast,kim2025test,ko2026active,hu2026turning}
has therefore attracted increasing attention, allowing a pretrained policy to
adapt online from observations and interactions encountered during deployment.
Such capability is important for bringing VLN from benchmark generalization
toward practical navigation in open-world environments.

The difficulty of TTA-VLN stems from the mismatch between training and
deployment distributions in both vision and language.
Unseen environments may introduce new appearances, layouts, and local
navigation contexts, while unseen instructions may contain unfamiliar
expressions.
Such shifts can make the pretrained policy's learned cross-modal decision
preferences unreliable, leading to off-course behavior.
Since the required correction is not explicitly available at test time,
existing TTA-VLN methods infer how to adapt from predictive uncertainty,
trajectory-level feedback, or accumulated adaptation experience.
However, these signals do not directly evaluate the actual consequence of the
current executed action, providing limited evidence of whether that interaction
truly contributes to instruction-guided progress.
Moreover, even when a signal suggests a plausible correction, the resulting
policy change may still be unreliable and should not necessarily persist.

\begin{figure*}[t]
  \centering
  \includegraphics[width=\textwidth]{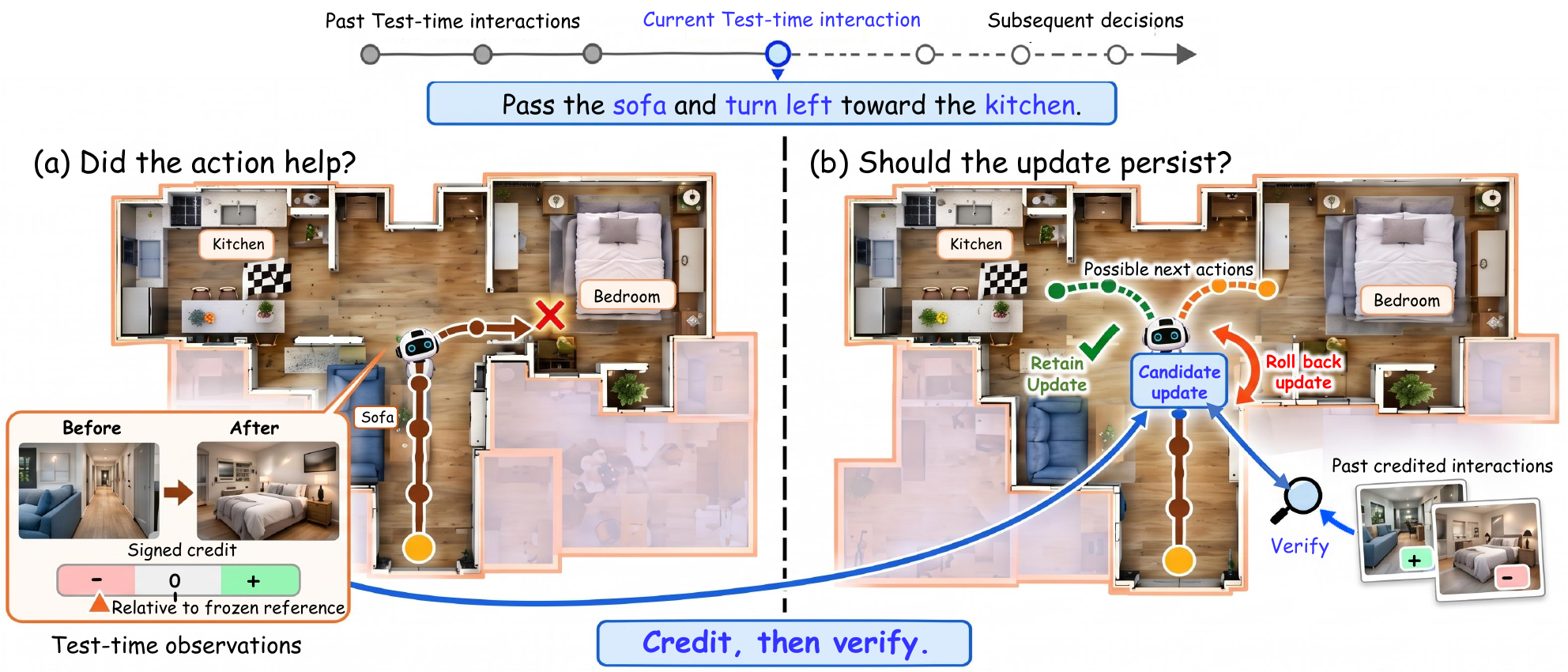}
  \caption{
  \textbf{Motivation for \method.}
  TTA enables online policy change, but change does not necessarily imply
improvement.
Decision credit first identifies whether the current interaction provides
local evidence of goal-directed improvement; the resulting candidate update
is then verified against past credit-supported interactions before being
retained or rolled back.
  }
  \vspace{-15pt}
  \label{fig:moti}
\end{figure*}

The key issue is therefore that TTA enables \emph{policy change}, but policy
change does not necessarily imply \emph{policy improvement}.
As illustrated in Fig.~\ref{fig:moti}, reliable TTA-VLN requires answering
two coupled questions:
\emph{how to identify whether the current interaction supports goal-directed
improvement, and how to determine whether the resulting candidate policy
update is worth retaining.} We refer to these two challenges as \emph{improvement identification} and
\emph{improvement retention}, respectively.
We observe that every executed action naturally produces an immediate
observation transition that exposes its local consequence.
This provides interaction-grounded evidence for identifying a supported
direction of change, while the resulting policy modification should still be
treated as a candidate that must be verified before it is allowed to persist.

Based on this insight, we propose Credit-Guided Policy Improvement
(CGPI), a framework for TTA-VLN that explicitly separates
\emph{improvement identification} from \emph{improvement retention}.
For improvement identification, CGPI grounds each executed action with its
actual resulting observation and compares the observed consequence against
the frozen pretrained policy's local reference expectation, yielding signed,
reference-relative decision credit without external outcome feedback.
The recovered credit provides directional guidance for a lightweight action
reranker, producing a candidate policy update rather than immediately
committing the change.
For improvement retention, CGPI verifies the candidate against previously
observed credit-supported interactions and retains it only when the
modification remains sufficiently supported and satisfies the proposal-state
reference constraint; otherwise, the policy rolls back to the last accepted
state.
The pretrained backbone remains frozen throughout adaptation, forming an
\emph{identify--propose--verify--retain} loop for continual test-time policy
improvement.
Experiments across three VLN benchmarks and four navigation backbones evaluate
CGPI, while qualitative trials on a physical robot further illustrate sim-to-real feasibility.

Our contributions are threefold.
\textit{\textbf{First}}, we formulate reliable TTA-VLN from the perspective of
\emph{policy improvement} rather than policy change, identifying two coupled
problems: \emph{improvement identification} and \emph{improvement retention}.
\textit{\textbf{Second}}, we propose Credit-Guided Policy Improvement
(CGPI), which recovers signed, reference-relative decision credit from
action-induced observation transitions and verifies credit-guided candidate
updates before selectively retaining or rolling them back.
\textit{\textbf{Third}}, experiments across REVERIE, R2R, and R2R-CE with
HAMT, DUET, BEVBert, and ETPNav evaluate CGPI across discrete and continuous
VLN settings; qualitative trials on a Unitree Go2 further illustrate
zero-shot sim-to-real feasibility.

\section{Method}

\subsection{Problem Setup and Overview}
\label{sec:problem}
\noindent
We consider vision-language navigation in a streaming test-time setting.
The agent is given a pretrained policy $\pi_{\theta_0}$ and receives a
stream of navigation tasks $\mathcal{X}=\{X_1,\ldots,X_N\}$.
Each task $X_i=(I_i,o_0^i)$ contains a natural-language instruction $I_i$
and an initial observation $o_0^i$.
At step $t$, the state is
$s_t=(I_i,o_t,\tau_{<t},\mathcal{G}_t)$, where $\tau_{<t}$ denotes the
navigation history and $\mathcal{G}_t$ denotes the currently available
local candidate structure.
The agent selects $a_t\in\mathcal{A}_t$ until it chooses $\stopact$.

We study \emph{feedback-free} test-time adaptation, by which we mean that
no external task-level correctness or outcome signal is available during
deployment.
The agent cannot access expert actions, target locations, dense rewards,
trajectory-level success labels, or human feedback.
For a movement action, the only new environment information obtained after
execution is its resulting observation $o_{t+1}$.
Candidate evidence for unexecuted actions is constructed only from
information already observable at $s_t$; no future observation of an
unexecuted action is accessed.

The frozen backbone produces base logits $z_t^0(a)$ and the reference
policy $\pi_0(a|s_t)$.
We keep all backbone parameters fixed and adapt only a lightweight residual
reranker $r_\phi$.
Let $\phi_t$ denote the currently retained reranker parameters.
The adapted policy is
\begin{equation}
\tilde z_t(a;\phi_t)
=
z_t^0(a)+\alpha\tanh\!\left(r_{\phi_t}(s_t,a)\right),
\qquad
\tilde\pi_{\phi_t}(a|s_t)
=
\mathrm{softmax}_{a\in\mathcal{A}_t}
\!\left(\tilde z_t(a;\phi_t)\right).
\label{eq:adapted_policy}
\end{equation}
Here, $\alpha$ bounds the maximum residual-logit correction.

Our goal is to identify interaction-induced policy changes that are locally
supported and to determine which resulting modifications should persist.
We denote a candidate modification proposed from the current interaction by
$\phi_t^{+}$.
A candidate is committed only after verification; otherwise, the reranker
remains at the previously retained state $\phi_t$.

\begin{figure*}[t]
  \centering
  \includegraphics[width=\textwidth]{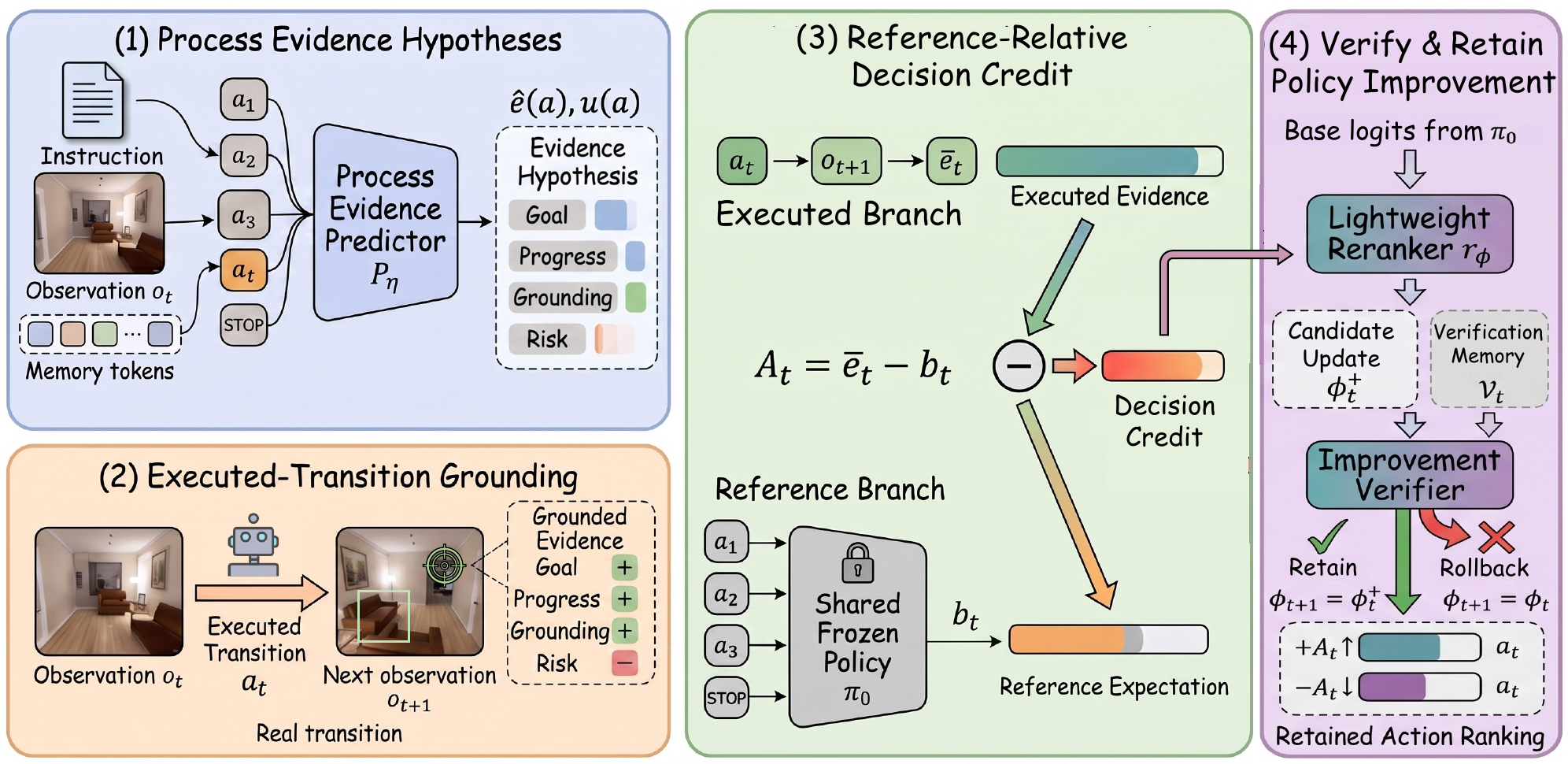}
  \caption{
  Overview of \method{}.
  The method constructs local process-evidence hypotheses and grounds the
  executed decision using its actual resulting observation.
  Reference-relative comparison recovers signed decision credit, which
  proposes a candidate modification to an action reranker.
  The candidate is then verified against past credit-supported interactions
and retained only when sufficiently supported and satisfying the
proposal-state reference constraint; otherwise, it is rolled back.
  }
  \vspace{-15pt}
  \label{fig:1}
\end{figure*}

As illustrated in Fig.~\ref{fig:1}, \method{} follows an
\emph{identify--propose--verify--retain} loop.
First, the consequence of an executed interaction is grounded by the actual
next observation and evaluated relative to the frozen policy's local
expectation, yielding reference-relative decision credit.
Second, sufficiently supported credit proposes a candidate modification to
the lightweight reranker.
Third, the candidate is evaluated against previously observed
credit-supported interactions.
Only a verified candidate is retained; otherwise, \method{} rolls back to
the last accepted reranker state.
The pretrained VLN backbone remains frozen throughout this process.

\subsection{Interaction-Grounded Improvement Identification}
\label{sec:identification}
\label{sec:reference_credit}

For movement-action credit, we define the movement candidate set as
$\mathcal{A}_t^{\mathrm{mov}}
=
\mathcal{A}_t\setminus\{\stopact\}$.
We renormalize the frozen policy over movement actions as
$\pi_0^{\mathrm{mov}}(a|s_t)
=
\pi_0(a|s_t)/
\sum_{a'\in\mathcal{A}_t^{\mathrm{mov}}}
\pi_0(a'|s_t)$,
so that
$\sum_{a\in\mathcal{A}_t^{\mathrm{mov}}}
\pi_0^{\mathrm{mov}}(a|s_t)=1$.
This renormalization is used only to construct the movement-action
reference; the original policy over $\mathcal{A}_t$, including
$\stopact$, is still used for navigation and action selection.

For each evidence view $k$, we construct a current-state hypothesis for
every locally available movement action.
After executing a non-STOP action $a_t$, the actual next observation
$o_{t+1}$ grounds the evidence for the executed decision:
\begin{equation}
\hat e_t^k(a)=\hat E^k(s_t,a),
\quad a\in\mathcal{A}_t^{\mathrm{mov}},
\qquad
e_t^k=E^k(s_t,a_t,o_{t+1}),
\quad a_t\neq\stopact .
\label{eq:evidence_pair}
\end{equation}
Here, $\hat E^k$ is computed entirely from information available at
$s_t$, while $e_t^k$ is grounded by the actual observation obtained after
executing $a_t$.
We instantiate four lightweight evidence terms:
\textbf{Goal}, \textbf{Progress}, \textbf{Grounding}, and \textbf{Risk};
we instantiate these four terms using lightweight evidence functions.

Because different evidence terms have different numerical scales, we
standardize them using source-training statistics computed once and fixed
throughout deployment:
\begin{equation}
\bar e_t^k
=
\frac{e_t^k-\mu_k}{\sigma_k+\epsilon},
\qquad
\bar{\hat e}_t^k(a)
=
\frac{\hat e_t^k(a)-\mu_k}{\sigma_k+\epsilon}.
\label{eq:normalization}
\end{equation}
No target labels, target locations, or test-time outcome
signals are used to construct these statistics.

Observation grounding indicates what happened after the executed action,
but an absolute evidence value does not reveal whether that consequence
is locally favorable relative to the movement alternatives available at
the same state.
We therefore compare the grounded consequence against the frozen
policy's movement-only local reference:
\begin{equation}
b_t^k
=
\sum_{a\in\mathcal{A}_t^{\mathrm{mov}}}
\pi_0^{\mathrm{mov}}(a|s_t)\,
\bar{\hat e}_t^k(a),
\qquad
C_t^k
=
\bar e_t^k-b_t^k,
\qquad
A_t
=
\frac{1}{K}\sum_{k=1}^{K}C_t^k,
\;\; a_t\neq\stopact .
\label{eq:decision_credit}
\end{equation}
Here, $b_t^k$ is the frozen-policy local reference over movement actions,
$C_t^k$ is the view-wise reference-relative credit, and $A_t$ is the
final movement-action credit.

The resulting $A_t$ is a \emph{local, reference-relative} estimate of the
executed action's consequence.
Positive credit indicates that the observed consequence is more favorable
than the frozen policy's movement-only local reference expectation under
the chosen evidence functions, while negative credit indicates the
opposite.
Since $\stopact$ does not induce a new post-action observation, it is not
included in the movement-action reference above.
We handle STOP separately using a conservative
completion-versus-movement credit, using a conservative completion-versus-movement credit.

\subsection{Credit-Guided Candidate Policy Update}
\label{sec:candidate_improvement}
\label{sec:credit_update}

Decision credit provides the direction for a candidate policy change.
We verify this candidate before commitment, separating
\emph{update proposal} from \emph{update retention}.

Let
$D_t(\phi)=
\KL(\tilde\pi_{\phi}(\cdot|s_t)\|\pi_0(\cdot|s_t))$
denote deviation from the frozen navigation prior.
We clip extreme credits and propose an update only when the credit is
sufficiently strong and the retained policy has not already drifted
excessively:
\begin{equation}
\hat A_t=\operatorname{clip}(A_t,-A_{\max},A_{\max}),
\qquad
m_t^{\mathrm{prop}}
=
\mathbf{1}[|A_t|\geq\delta_A]\,
\mathbf{1}[D_t(\phi_t)\leq\delta_{\mathrm{KL}}].
\label{eq:proposal_gate}
\end{equation}
The credit-magnitude condition filters weak signals whose directions are most sensitive to evidence-estimation error. 

For an eligible interaction, the candidate-proposal objective is
\begin{equation}
\mathcal{L}_{\mathrm{prop}}(t)
=
-\sg(\hat A_t)\log\tilde\pi_{\phi_t}(a_t|s_t)
+\lambda_{\mathrm{KL}}D_t(\phi_t)
+\lambda_{\mathrm{reg}}\|\phi_t\|_2^2 .
\label{eq:proposal_loss}
\end{equation}
Positive credit therefore proposes increasing the corresponding decision
tendency, whereas negative credit proposes suppressing it.
The KL term regularizes the proposal toward the frozen navigation prior.
One optimizer step on $\mathcal{L}_{\mathrm{prop}}(t)$ produces candidate
model and optimizer states
$(\phi_t^{+},\omega_t^{+})
=\operatorname{OptStep}((\phi_t,\omega_t),
\mathcal{L}_{\mathrm{prop}}(t))$,
where $\omega_t$ is the optimizer state associated with the currently
retained reranker.
We pass $(\phi_t^{+},\omega_t^{+})$ to the verification stage as the
candidate model and optimizer states.

\subsection{Verify-and-Retain Policy Improvement}
\label{sec:verification}

To determine whether a candidate modification deserves to persist, we maintain
a bounded verification memory
$\mathcal{V}_t=\{(s_i,a_i,A_i)\mid i\in\mathcal{I}_t\}$
of recent credit-supported interactions, where
$|A_i|\geq\delta_A$ and $i<t$.
The current interaction that produces $\phi_t^{+}$ is excluded during its own
verification and inserted only after the retain-or-rollback decision,
preventing self-validation.

For any reranker parameters $\phi$, we measure consistency with stored
decision credits and define the candidate verification gain as
\begin{equation}
J_t(\phi)=
\frac{1}{|\mathcal{V}_t|}
\sum_{i\in\mathcal{I}_t}
\sg(A_i)\log\tilde\pi_{\phi}(a_i|s_i),
\qquad
\Delta_t^{\mathrm{ver}}
=
J_t(\phi_t^{+})-J_t(\phi_t).
\label{eq:verification}
\end{equation}
For positive stored credit, increasing the corresponding action probability
raises $J_t$; for negative stored credit, decreasing it also raises $J_t$.
Hence $\Delta_t^{\mathrm{ver}}$ measures whether the candidate is more
consistent with previously supported decision directions than the currently
retained policy.

We retain a candidate only when it improves the verification objective while
remaining sufficiently close to the frozen navigation prior:
\begin{equation}
r_t=
\mathbf{1}[\Delta_t^{\mathrm{ver}}\geq\delta_{\mathrm{keep}}]\,
\mathbf{1}[D_t(\phi_t^{+})\leq\delta_{\mathrm{KL}}].
\label{eq:retention}
\end{equation}
If $r_t=1$, both candidate states
$(\phi_t^{+},\omega_t^{+})$ are retained;
otherwise, the policy and optimizer remain at
$(\phi_t,\omega_t)$.
Thus, decision credit determines what change should be proposed, whereas
verification determines whether that change should persist.

If no candidate is proposed or the verification memory has not reached its
warm-up size, the retained state remains unchanged.
After the retain-or-rollback decision, an eligible current interaction is
inserted into the bounded verification memory and becomes available for
evaluating future candidates.

\begin{table*}[b]
\caption{Experimental results for different TTA strategies on the REVERIE dataset.}
\label{tab:main_reverie}
\centering
\small
\setlength{\tabcolsep}{3.2pt}
\renewcommand{\arraystretch}{0.96}
\resizebox{\textwidth}{!}{%
\begin{tabular}{l|cccc|cccc|cccc}
\toprule
\multirow{2}{*}{\textbf{Methods+Model}}
& \multicolumn{4}{c|}{\textbf{REVERIE Val Seen}}
& \multicolumn{4}{c|}{\textbf{REVERIE Val Unseen}}
& \multicolumn{4}{c}{\textbf{REVERIE Test Unseen}} \\
& OSR$\uparrow$ & SR$\uparrow$ & SPL$\uparrow$ & RGSPL$\uparrow$
& OSR$\uparrow$ & SR$\uparrow$ & SPL$\uparrow$ & RGSPL$\uparrow$
& OSR$\uparrow$ & SR$\uparrow$ & SPL$\uparrow$ & RGSPL$\uparrow$ \\
\midrule
HAMT~\citep{chen2021history}
& 47.65 & 43.29 & 40.19 & 25.18 & 36.84 & 32.95 & 30.20 & 17.28 & 33.41 & 30.40 & 26.67 & 13.08 \\
\hspace{0.7em}+ Tent~\citep{wang2020tent}
& 46.03 & 43.43 & 40.78 & 25.81 & 32.60 & 30.56 & 28.23 & 14.48 & 25.06 & 23.73 & 21.78 & 10.82 \\
\hspace{0.7em}+ SAR~\citep{niu2023towards}
& 47.12 & 43.85 & 39.97 & 25.44 & 32.86 & 31.12 & 29.15 & 15.23 & 27.94 & 25.86 & 23.08 & 11.58 \\
\hspace{0.7em}+ ViDA~\citep{liu2024vida}
& 47.67 & 43.55 & 41.23 & 25.27 & 32.74 & 30.97 & 28.82 & 14.97 & 27.03 & 24.81 & 22.45 & 11.23 \\
\hspace{0.7em}+ FSTTA~\citep{gao2024fast}
& 48.21 & 42.87 & 39.56 & 24.58 & 36.78 & 32.89 & 30.51 & 17.20 & 33.39 & 30.39 & 26.65 & 13.61 \\
\hspace{0.7em}+ ReCAP~\citep{hu2025beyond}
& 48.49 & 44.06 & 40.69 & 25.46 & 37.04 & 33.06 & 30.28 & 17.37 & 34.11 & 30.51 & 24.27 & 13.11 \\
\hspace{0.7em}+ IDEA~\citep{hu2026turning}
& 50.67 & 47.33 & 42.13 & 26.82 & 39.87 & 34.92 & 31.52 & 17.76 & 38.14 & 32.81 & 28.52 & 14.45 \\
\rowcolor{resultblue}
\hspace{0.7em}+ \textbf{Ours}
& \textbf{53.01} & \textbf{49.19} & \textbf{44.22} & \textbf{28.78}
& \textbf{42.05} & \textbf{36.97} & \textbf{33.26} & \textbf{19.98}
& \textbf{40.25} & \textbf{34.73} & \textbf{30.78} & \textbf{15.48} \\
\midrule
DUET~\citep{chen2022think}
& 73.86 & 71.75 & 63.94 & 51.14 & 51.07 & 46.98 & 33.73 & 23.03 & 56.91 & 52.51 & 36.06 & 22.06 \\
\hspace{0.7em}+ Tent~\citep{wang2020tent}
& 73.72 & 71.89 & 64.06 & 50.41 & 51.43 & 47.55 & 33.99 & 23.32 & 57.12 & 52.61 & 36.17 & 22.16 \\
\hspace{0.7em}+ SAR~\citep{niu2023towards}
& 74.84 & 71.75 & 64.43 & 51.70 & 53.26 & 48.00 & 33.92 & 23.09 & 57.11 & 53.04 & 36.07 & 22.27 \\
\hspace{0.7em}+ ViDA~\citep{liu2024vida}
& 73.99 & 72.49 & 63.49 & 50.89 & 52.53 & 48.14 & 32.45 & 21.92 & 56.78 & 52.74 & 35.10 & 21.77 \\
\hspace{0.7em}+ FSTTA~\citep{gao2024fast}
& 75.59 & 75.48 & 65.84 & 52.23 & 56.26 & 54.15 & 36.41 & 23.56 & 58.44 & 53.40 & 36.43 & 22.40 \\
\hspace{0.7em}+ ReCAP~\citep{hu2025beyond}
& 75.06 & 74.72 & 65.87 & 53.42 & 56.67 & 54.74 & 36.22 & 23.74 & 57.72 & 53.07 & 36.52 & 22.47 \\
\hspace{0.7em}+ IDEA~\citep{hu2026turning}
& 78.45 & 78.24 & 67.74 & 55.07 & 58.51 & 56.92 & 38.03 & 25.47 & 58.91 & 55.12 & 39.84 & 24.52 \\
\rowcolor{resultblue}
\hspace{0.7em}+ \textbf{Ours}
& \textbf{79.62} & \textbf{79.22} & \textbf{68.05} & \textbf{56.96}
& \textbf{60.75} & \textbf{58.76} & \textbf{40.15} & \textbf{27.54}
& \textbf{60.88} & \textbf{57.31} & \textbf{41.87} & \textbf{24.67} \\
\bottomrule
\end{tabular}%
}
\end{table*}

\section{Experiments}

\paragraph{Experimental setup.}
We evaluate CGPI on REVERIE~\citep{qi2020reverie},
R2R~\citep{anderson2018vision}, and R2R-CE~\citep{krantz2020beyond}
with four pretrained navigation policies:
HAMT~\citep{chen2021history}, DUET~\citep{chen2022think},
BEVBert~\citep{an2023bevbert}, and ETPNav~\citep{an2024etpnav}.
Following prior TTA-VLN evaluation protocols~\citep{hu2026turning},
we use the standard benchmark splits, pretrained policies, and official
episode order under continual test-time adaptation.

\subsection{Main Results}

\paragraph{Evaluation on REVERIE.}
Table~\ref{tab:main_reverie} compares \method with existing TTA methods on REVERIE using HAMT and DUET as navigation backbones. \method consistently achieves the best performance across Val Seen, Val Unseen, and Test Unseen. On DUET, \method improves SR/SPL from 56.92/38.03 to 58.76/40.15 on Val Unseen and from 55.12/39.84 to 57.31/41.87 on Test Unseen compared with IDEA. Similar improvements are observed with HAMT, where SR/SPL increases from 34.92/31.52 to 36.97/33.26 on Val Unseen. These results show that combining observation-grounded,
reference-relative improvement identification with selective
verify-and-retain adaptation provides an effective test-time policy
improvement mechanism across navigation backbones.

\begin{table*}[t]
\centering
\small
\setlength{\tabcolsep}{3pt}
\renewcommand{\arraystretch}{0.96}
\begin{minipage}[t]{0.475\textwidth}
\centering
\caption{Experiments on the R2R dataset.}
\label{tab:main_r2r}
\resizebox{\linewidth}{!}{%
\begin{tabular}{l|cccc|cccc}
\toprule
\multirow{2}{*}{\textbf{Methods}}
& \multicolumn{4}{c|}{\textbf{R2R Val Seen}}
& \multicolumn{4}{c}{\textbf{R2R Val Unseen}} \\
& TL$\downarrow$ & NE$\downarrow$ & SR$\uparrow$ & SPL$\uparrow$
& TL$\downarrow$ & NE$\downarrow$ & SR$\uparrow$ & SPL$\uparrow$ \\
\midrule
DUET~\citep{chen2022think} & 12.33 & 2.28 & 79 & 73 & 13.94 & 3.31 & 72 & 60 \\
\hspace{0.7em}+ Tent & 12.17 & 2.38 & 78 & 72 & 13.78 & 3.42 & 72 & 60 \\
\hspace{0.7em}+ SAR & 12.05 & 2.28 & 78 & 72 & 13.59 & 3.28 & 72 & 61 \\
\hspace{0.7em}+ ViDA & 12.11 & 2.31 & 79 & 73 & 13.63 & 3.34 & 72 & 61 \\
\hspace{0.7em}+ FSTTA & 13.39 & 2.25 & 79 & 73 & 14.64 & 3.03 & 75 & 62 \\
\hspace{0.7em}+ ReCAP & 12.02 & 2.27 & 78 & 73 & 13.39 & 3.28 & 72 & 61 \\
\hspace{0.7em}+ IDEA & 11.23 & 2.03 & 81 & 76 & 12.47 & 2.91 & 76 & 67 \\
\rowcolor{resultblue}
\hspace{0.7em}+ \textbf{Ours} & \textbf{10.91} & \textbf{1.89} & \textbf{83} & \textbf{77} & \textbf{12.08} & \textbf{2.73} & \textbf{79} & \textbf{68} \\
\midrule
BEVBert~\citep{an2023bevbert} & 13.56 & 2.17 & 81 & 74 & 14.55 & 2.81 & 75 & 64 \\
\hspace{0.7em}+ Tent & 12.68 & 2.36 & 80 & 74 & 13.14 & 2.93 & 74 & 63 \\
\hspace{0.7em}+ SAR & 12.49 & 2.28 & 80 & 74 & 12.98 & 2.79 & 75 & 64 \\
\hspace{0.7em}+ ViDA & 12.53 & 2.31 & 81 & 74 & 13.11 & 2.83 & 75 & 64 \\
\hspace{0.7em}+ FSTTA & 12.28 & 2.31 & 80 & 75 & 13.96 & 2.89 & 74 & 63 \\
\hspace{0.7em}+ ReCAP & 12.31 & 2.27 & 81 & 74 & 12.94 & 2.78 & 75 & 64 \\
\hspace{0.7em}+ IDEA & 10.89 & 2.15 & 83 & 79 & 12.03 & 2.53 & 76 & 68 \\
\rowcolor{resultblue}
\hspace{0.7em}+ \textbf{Ours} & \textbf{10.61} & \textbf{2.02} & \textbf{84} & \textbf{81} & \textbf{11.69} & \textbf{2.38} & \textbf{79} & \textbf{70} \\
\bottomrule
\end{tabular}%
}
\end{minipage}
\hfill
\begin{minipage}[t]{0.50\textwidth}
\centering
\caption{Experiments on the R2R-CE dataset.}
\label{tab:main_r2rce}
\resizebox{\linewidth}{!}{%
\begin{tabular}{l|ccccc|ccccc}
\toprule
\multirow{2}{*}{\textbf{Methods}}
& \multicolumn{5}{c|}{\textbf{R2R-CE Val Seen}}
& \multicolumn{5}{c}{\textbf{R2R-CE Val Unseen}} \\
& TL$\downarrow$ & NE$\downarrow$ & OSR$\uparrow$ & SR$\uparrow$ & SPL$\uparrow$
& TL$\downarrow$ & NE$\downarrow$ & OSR$\uparrow$ & SR$\uparrow$ & SPL$\uparrow$ \\
\midrule
BEVBert~\citep{an2023bevbert} & 13.98 & 3.77 & 73 & 68 & 60 & 13.27 & 4.57 & 67 & 59 & 50 \\
\hspace{0.7em}+ Tent & 12.74 & 3.35 & 76 & 70 & 62 & 13.29 & 4.61 & 65 & 59 & 49 \\
\hspace{0.7em}+ SAR & 12.53 & 3.28 & 76 & 70 & 63 & 13.23 & 4.54 & 66 & 59 & 50 \\
\hspace{0.7em}+ ViDA & 12.61 & 3.31 & 76 & 71 & 62 & 13.26 & 4.57 & 67 & 59 & 49 \\
\hspace{0.7em}+ FSTTA & 14.07 & 4.11 & 74 & 69 & 60 & 13.11 & 4.39 & 65 & 60 & 51 \\
\hspace{0.7em}+ ReCAP & 12.40 & 3.31 & 76 & 71 & 63 & 13.01 & 4.57 & 66 & 60 & 50 \\
\hspace{0.7em}+ IDEA & 12.27 & 3.04 & 78 & 73 & 64 & 12.67 & 4.26 & 69 & 62 & 52 \\
\rowcolor{resultblue}
\hspace{0.7em}+ \textbf{Ours} & \textbf{11.93} & \textbf{2.92} & \textbf{81} & \textbf{75} & \textbf{65} & \textbf{12.33} & \textbf{4.13} & \textbf{70} & \textbf{65} & \textbf{54} \\
\midrule
ETPNav~\citep{an2024etpnav} & 11.78 & 3.95 & 72 & 66 & 59 & 11.99 & 4.71 & 65 & 57 & 49 \\
\hspace{0.7em}+ Tent & 11.36 & 3.94 & 72 & 66 & 59 & 11.57 & 4.74 & 64 & 57 & 49 \\
\hspace{0.7em}+ SAR & 11.31 & 3.89 & 72 & 66 & 60 & 11.55 & 4.67 & 65 & 57 & 49 \\
\hspace{0.7em}+ ViDA & 11.58 & 3.91 & 72 & 66 & 60 & 11.62 & 4.72 & 64 & 57 & 49 \\
\hspace{0.7em}+ FSTTA & 11.35 & 3.93 & 72 & 66 & 59 & 11.57 & 4.77 & 64 & 57 & 49 \\
\hspace{0.7em}+ ReCAP & 11.31 & 3.92 & 72 & 66 & 60 & 11.56 & 4.74 & 65 & 57 & 50 \\
\hspace{0.7em}+ IDEA & 10.87 & 3.82 & 73 & 68 & 62 & 11.17 & 4.47 & 67 & 59 & 51 \\
\rowcolor{resultblue}
\hspace{0.7em}+ \textbf{Ours} & \textbf{10.59} & \textbf{3.70} & \textbf{75} & \textbf{69} & \textbf{65} & \textbf{10.90} & \textbf{4.34} & \textbf{70} & \textbf{61} & \textbf{52} \\
\bottomrule
\end{tabular}%
}
\end{minipage}
\end{table*}

\paragraph{Evaluation on R2R \& R2R-CE.}
Tables~\ref{tab:main_r2r} and~\ref{tab:main_r2rce} further evaluate \method on instruction-following navigation in both discrete and continuous environments. On R2R Val Unseen, \method improves SR/SPL from 76/67 to 79/68 with DUET and from 76/68 to 79/70 with BEVBert compared with IDEA. On R2R-CE Val Unseen, \method also improves SR/SPL from 62/52 to 65/54 with BEVBert and from 59/51 to 61/52 with ETPNav. These results establish gains across the four evaluated navigation architectures in both discrete and continuous VLN settings.

\begin{figure*}[t]
  \centering
  \includegraphics[width=\textwidth]{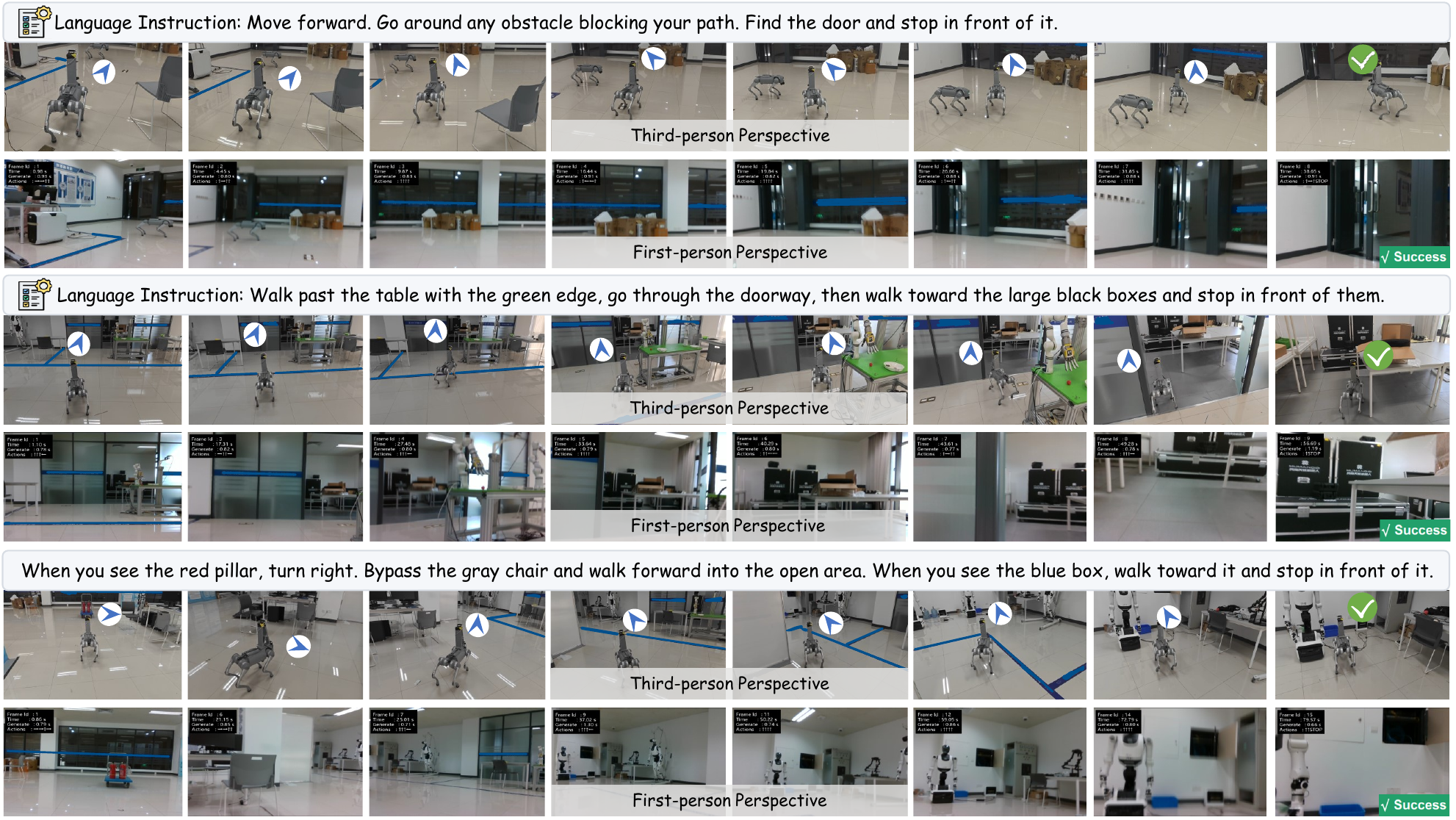}
\caption{Real-world qualitative results of \method on a Unitree Go2 robot. We show third-person views and egocentric observations for three executed
trajectories.}
  \label{fig:real}
\end{figure*}

\paragraph{Real-world qualitative results.}
We further deploy CGPI on a Unitree Go2 using a navigation policy trained
only in simulated environments, without real-world fine-tuning
or external outcome feedback.
The three trajectories in Fig.~\ref{fig:real} illustrate successful
multi-stage instruction following and sim-to-real feasibility in unseen environments. 

\subsection{Case Studies of Credit-Guided Policy Improvement}
\label{sec:case_study}

\begin{figure}[t]
    \centering

    \begin{subfigure}{0.495\linewidth}
        \centering
        \includegraphics[width=\linewidth]{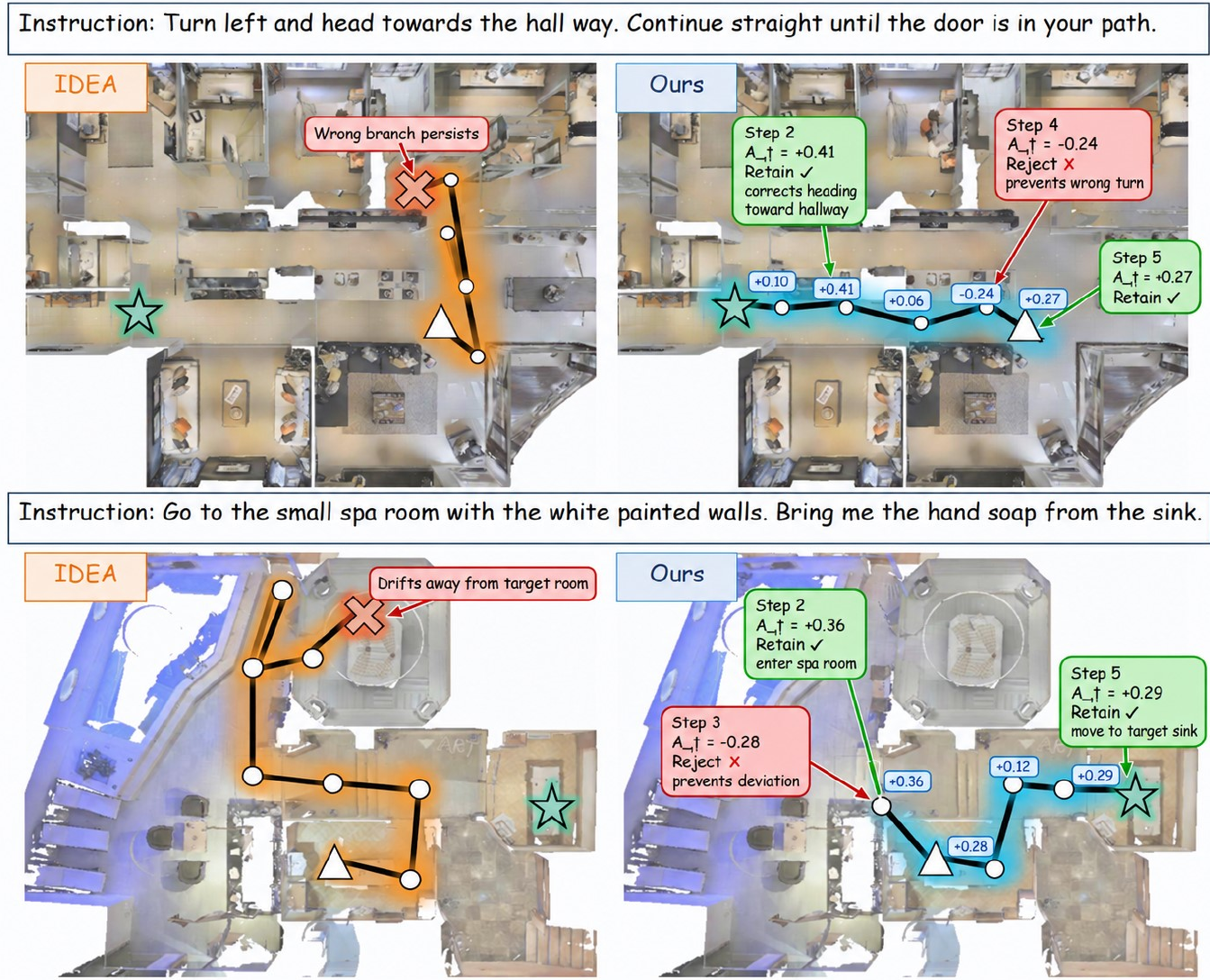}
    \end{subfigure}
    \hfill
    \begin{subfigure}{0.495\linewidth}
        \centering
        \includegraphics[width=\linewidth]{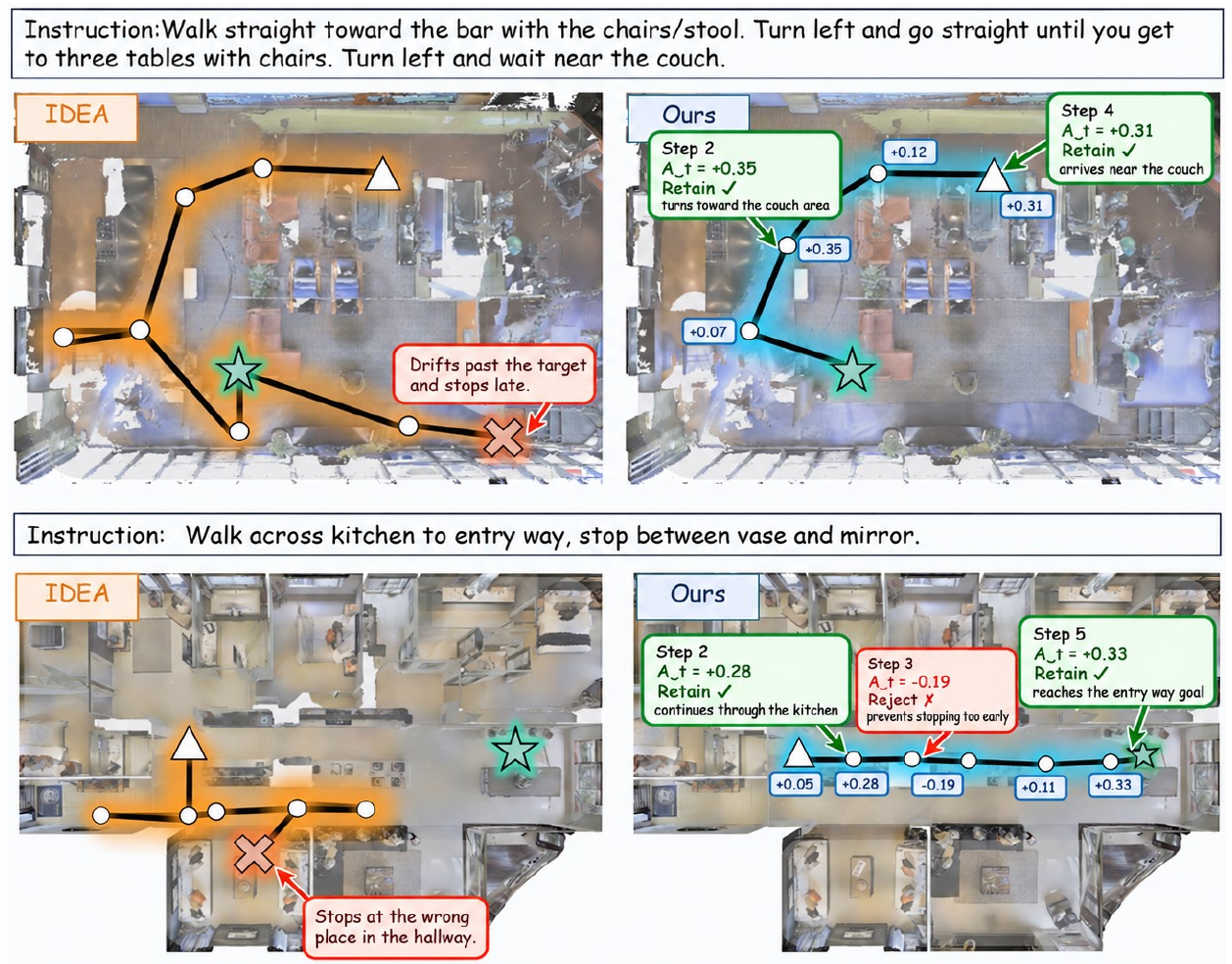}
    \end{subfigure}
    \caption{
Qualitative comparison of IDEA and CGPI in unseen environments.
}
    \label{fig:sim_qualitative_sup}
\end{figure}

Figure~\ref{fig:sim_qualitative_sup} visualizes representative test-time policy improvements in unseen environments, where \method{} trajectories mark key decision points with recovered decision credit $A_t$ and associated verification outcomes.
For the instruction
    \emph{``Go to the small spa room with the white painted walls. Bring me
    the hand soap from the sink,''}
    \method{} first obtains a positive credit $A_t=+0.36$; the resulting
    candidate update is retained and shifts the policy toward the spa room.
    At the following decision point, a negative credit $A_t=-0.28$ produces
    a candidate update that is rejected by verification, preventing the
    policy from drifting toward the wrong side room.
    A later positive credit $A_t=+0.29$ is retained, further reinforcing
    movement toward the target sink.
    In contrast, IDEA continues along an off-course branch and stops away
    from the target.
This visualizes how decision credit determines the update direction and
verification controls whether that update is retained.

\subsection{Ablation of Improvement Identification}
\label{sec:core_ablation}

We isolate the two operations used for improvement identification:
\emph{observation grounding}, which evaluates the executed action using
its actual next observation, and \emph{reference-relative comparison},
which evaluates the resulting evidence relative to the frozen policy's
local expectation.
All other CGPI components, including candidate proposal, verification,
and retention, are kept fixed.

For this ablation, we use the view-averaged grounded evidence
$\bar e_t=\frac{1}{K}\sum_k\bar e_t^k$, the view-averaged pre-action
hypothesis
$\bar{\hat e}_t(a)=\frac{1}{K}\sum_k\bar{\hat e}_t^k(a)$, and the
view-averaged reference
$b_t=\frac{1}{K}\sum_k b_t^k$.
The four variants differ only in the signal used for candidate proposal:
\textbf{Absolute Hypothesis} uses $\bar{\hat e}_t(a_t)$;
\textbf{Grounded Evidence} uses $\bar e_t$;
\textbf{Relative Hypothesis} uses
$\bar{\hat e}_t(a_t)-b_t$; and
\textbf{CGPI} uses $\bar e_t-b_t$, which is equivalent to
Eq.~\ref{eq:decision_credit} for movement actions.

To assess the quality of the proposal signal, we additionally report two
post-hoc diagnostics.
\textbf{Credit--Progress Correlation (CPC)} measures the Spearman
correlation between the test-time signal and actual one-step geodesic
progress, while \textbf{Wrong Update Rate (WUR)} measures the fraction of
proposal-triggering interactions whose signal direction disagrees with
oracle local progress.
Geodesic information is used only for post-hoc evaluation and is never available during adaptation.

\begin{table}[t]
    \centering
    \caption{
    Ablation of \method{}'s two core improvement-identification operations on REVERIE Val Unseen with DUET. 
    }
    \label{tab:core_ablation}
    \small
    \setlength{\tabcolsep}{5.5pt}
    \renewcommand{\arraystretch}{1.10}
    \resizebox{0.7\textwidth}{!}{%
    \begin{tabular}{l|cc|cccc}
        \toprule
        \textbf{Variant}
        & \textbf{Grounding}
        & \textbf{Reference}
        & SR$\uparrow$
        & SPL$\uparrow$
        & CPC$\uparrow$
        & WUR$\downarrow$ \\
        \midrule

        Absolute Hypothesis
        & -- & --
        & 54.32 & 36.71 & 0.297 & 0.312 \\

        Grounded Evidence
        & \checkmark & --
        & 56.14 & 38.06 & 0.414 & 0.244 \\

        Relative Hypothesis
        & -- & \checkmark
        & 56.82 & 38.67 & 0.463 & 0.221 \\

        \rowcolor{resultblue}
        \textbf{CGPI}
        & \checkmark & \checkmark
        & \textbf{58.76}
        & \textbf{40.15}
        & \textbf{0.572}
        & \textbf{0.154} \\
        \bottomrule
    \end{tabular}%
    }
\end{table}

Table~\ref{tab:core_ablation} shows complementary contributions from both
operations.
Observation grounding improves SR, SPL, and CPC over the pre-action
hypothesis, while reference-relative comparison further improves the
quality of the proposal signal.
Combining both yields the highest navigation performance and CPC together
with the lowest WUR.
This supports using the observed transition to determine what happened and
the frozen-policy reference to determine whether that consequence is locally
favorable.

\paragraph{Contribution of STOP credit.}
The core ablation above keeps STOP handling fixed in order to isolate
the two operations used for movement-action credit. We therefore further
separate the contributions of observation-grounded movement credit and
the dedicated STOP credit in Eq.~\ref{eq:stop_credit}. When movement credit is disabled, movement actions do not generate
credit-guided candidate policy modifications.
When STOP credit is disabled, STOP remains a valid action of the
navigation policy, but a STOP decision does not generate a candidate
modification.
All other CGPI components, including verification and retention, remain
unchanged.

\begin{table}[t]
    \centering
    \caption{
    Contribution of movement-action credit and STOP-specific credit in
\method{} on REVERIE.
    }
    \label{tab:stop_ablation}
    \small
    \setlength{\tabcolsep}{8pt}
    \renewcommand{\arraystretch}{1.08}
\begin{tabular}{cc|cc}
    \toprule
    \textbf{Movement Credit}
    & \textbf{STOP Credit}
    & \textbf{SR$\uparrow$}
    & \textbf{SPL$\uparrow$} \\
    \midrule

    -- & --
    & 30.40 & 26.67 \\

    \checkmark & --
    & 33.54 & 29.61 \\

    -- & \checkmark
    & 31.46 & 27.55 \\

    \rowcolor{resultblue}
    \checkmark & \checkmark
    & \textbf{34.73}
    & \textbf{30.78} \\

    \bottomrule
\end{tabular}
\end{table}
As shown in Table~\ref{tab:stop_ablation}, enabling movement-action credit alone 
improves SR and SPL from 30.40/26.67 to 33.54/29.61, accounting
for most of the gain over the frozen HAMT policy. Using only STOP credit also improves performance to
31.46/27.55, showing that learning from stopping decisions provides an
additional adaptation signal. Combining movement and STOP credit yields
the strongest result of 34.73 SR and 30.78 SPL, indicating that the two
sources of credit are complementary.
Because STOP terminates the current navigation episode, a retained update
induced by STOP cannot affect later decisions within the same trajectory.
Under the continual protocol, however, the retained policy modification
can influence subsequent episodes.

\subsection{Effect of Improvement Verification and Retention}
\label{sec:retention_ablation}

We next examine the second stage of \method{}: whether a candidate policy
change should be retained after it has been proposed by decision credit.
All variants use the same observation-grounded, reference-relative credit,
lightweight reranker, and candidate-proposal objective.
They differ only in how candidate modifications are evaluated before being
committed. Table~\ref{tab:retention_ablation} reports this ablation on REVERIE
Val Unseen with DUET.

We evaluate retention behavior using three complementary diagnostics.
\textbf{Policy Drift (PD)} measures the average KL divergence between the
adapted and frozen action distributions.
\textbf{Accepted Improvement Rate (AIR)} is the fraction of retained
movement-action candidates that yield positive short-horizon
counterfactual progress, while \textbf{Retention Rate (RR)} is the
fraction of proposed candidates that are committed.
These diagnostics are used only for analysis.

\begin{table}[H]
\centering
\caption{
Ablation of improvement verification and retention on REVERIE Val Unseen
with DUET.
}
\label{tab:retention_ablation}
\small
\setlength{\tabcolsep}{4.5pt}
\renewcommand{\arraystretch}{1.08}
\begin{tabular}{l|cc|ccccc}
\toprule
\textbf{Variant}
& \textbf{Verify}
& \textbf{Cand. KL}
& SR$\uparrow$
& SPL$\uparrow$
& PD$\downarrow$
& AIR$\uparrow$
& RR(\%) \\
\midrule

CGPI w/o Verification
& --
& --
& 57.41
& 38.96
& 0.071
& 0.548
& 100.0 \\

CGPI w/o Candidate KL
& \checkmark
& --
& 58.29
& 39.72
& 0.057
& 0.704
& 67.8 \\

\rowcolor{resultblue}
\textbf{Full CGPI}
& \checkmark
& \checkmark
& \textbf{58.76}
& \textbf{40.15}
& \textbf{0.043}
& \textbf{0.741}
& \textbf{61.3} \\

\bottomrule
\end{tabular}
\end{table}

The comparison isolates whether evaluating a proposed policy change
before committing it improves the quality of retained adaptations.
Without verification, every eligible candidate is committed, resulting
in a higher retention rate but also larger policy drift and a lower AIR.
Adding credit-based verification rejects unsupported modifications and
improves both navigation performance and retained-candidate quality.
The full method further applies the candidate-level KL condition,
reducing policy drift from 0.057 to 0.043 while increasing AIR from
0.704 to 0.741.

\subsection{Are Retained Policy Changes Actually Better?}
\label{sec:retention_quality}

To directly examine whether the verify-and-retain mechanism distinguishes
useful policy changes from harmful ones, we perform a post-hoc
counterfactual analysis on REVERIE Val Unseen with DUET.
For every movement-action candidate, we clone the complete navigation state
after the executed transition, including the simulator state, trajectory
history, and accumulated navigation structure.
The retained policy $\phi_t$ and candidate policy $\phi_t^{+}$ are then
rolled out independently from identical decision contexts for a short
horizon $H$.
Geodesic information is used only for this post-hoc diagnosis and is
never available to \method{} during adaptation.

Let $\Gamma_t^{(H)}$ denote the difference in short-horizon oracle
progress between the candidate and the currently retained policy.
We report:
(i) \textbf{Accepted Improvement Rate (AIR)}, the fraction of retained
candidates with $\Gamma_t^{(H)}>0$;
(ii) \textbf{Accepted Improvement Gain (AIG)}, the average
$\Gamma_t^{(H)}$ among retained candidates; and
(iii) \textbf{Rejection Precision (RP)}, the fraction of rejected
candidates with $\Gamma_t^{(H)}<0$.

As shown in Table~\ref{tab:retention_quality} (left), compared
with direct commitment, CGPI increases AIR from 0.548 to 0.741
and the mean accepted improvement gain from 0.19 to 0.63\,m.
Moreover, an RP of 0.716 indicates that 71.6\% of rejected candidates
would have produced negative short-horizon oracle progress under the
post-hoc counterfactual evaluation.
These results show that verification both improves the quality
of retained candidates and filters a substantial fraction of harmful policy changes.

\begin{table}[tbp]
\centering
\caption{
Policy-improvement analysis on REVERIE. Left: post-hoc candidate-update
quality (DUET, Val Unseen). Right: within-episode versus continual
adaptation (HAMT, Test Unseen).
}
\label{tab:retention_quality}
\label{tab:episode_reset}

\footnotesize
\setlength{\tabcolsep}{3.5pt}
\renewcommand{\arraystretch}{1.12}

\begin{minipage}[t]{0.54\linewidth}
\vspace{0pt}
\centering

\begin{tabularx}{\linewidth}{>{\raggedright\arraybackslash}Xccc}
\toprule
\textbf{Method}
& \textbf{AIR}$\uparrow$
& \textbf{AIG (m)}$\uparrow$
& \textbf{RP}$\uparrow$ \\
\midrule

Credit-Guided Direct Commit
& 0.548
& 0.19
& -- \\

\rowcolor{resultblue}
\textbf{CGPI}
& \textbf{0.741}
& \textbf{0.63}
& \textbf{0.716} \\

\bottomrule
\end{tabularx}
\end{minipage}%
\hfill
\begin{minipage}[t]{0.42\linewidth}
\vspace{0pt}
\centering

\begin{tabularx}{\linewidth}{>{\raggedright\arraybackslash}Xcc}
\toprule
\textbf{Setting}
& \textbf{SR}$\uparrow$
& \textbf{SPL}$\uparrow$ \\
\midrule

Frozen HAMT
& 30.40
& 26.67 \\

CGPI, episode reset
& 32.62
& 28.41 \\

\rowcolor{resultblue}
\textbf{CGPI, continual}
& \textbf{34.73}
& \textbf{30.78} \\

\bottomrule
\end{tabularx}
\end{minipage}

\end{table}



\subsection{Within-Episode versus Continual Adaptation}
\label{tab:episode_reset}

Since the decision credit for an executed action is obtained only after
observing its resulting transition, a natural question is whether the
recovered credit already benefits subsequent decisions within the same
episode or mainly contributes through accumulation across episodes.
We therefore compare \method{} under episode-reset and continual
adaptation protocols on REVERIE Test Unseen with HAMT.
In the \emph{episode-reset} setting, the reranker parameters, optimizer
state, and verification memory are restored to the same initialization
before every navigation episode.
All CGPI operations, including credit recovery, candidate proposal,
verification, and retention, are performed normally within the episode.
Thus, any improvement over the frozen policy can only arise from within-episode policy improvement.
In the \emph{continual} setting, the retained reranker parameters,
optimizer state, and verification memory are preserved across episodes,
following our default evaluation protocol.

As shown in Table~\ref{tab:episode_reset} (right), episode-reset
CGPI improves SR/SPL from 30.40/26.67 to 32.62/28.41 over
the frozen HAMT policy. Since the reranker and optimizer state are reset before every
episode, this improvement can only arise from CGPI updates affecting
subsequent decisions within the same trajectory. 
This isolates a clear within-episode adaptation benefit from post-action decision credit.
Retaining the adaptation state across episodes further improves
performance to 34.73/30.78, indicating an additional benefit from
cross-episode accumulation under the continual test-time setting.

\section{Conclusion}

We introduced \method{}, a feedback-free test-time adaptation framework
that formulates online VLN adaptation from the perspective of
\emph{policy improvement} rather than policy change.
CGPI uses action-induced observation transitions to recover local,
reference-relative decision credit and separates credit-guided update
proposal from subsequent verification and retention.
Across the evaluated discrete and continuous VLN settings, CGPI improves
navigation performance with multiple pretrained backbones while adapting
only a lightweight residual reranker.
Post-hoc analyses further show that the recovered credit is aligned with
local navigation progress and that selective verification improves the
quality of retained candidate updates.
Qualitative robot trials additionally illustrate zero-shot sim-to-real
feasibility in the shown real-world settings.

\bibliography{iclr2027_conference}
\bibliographystyle{iclr2027_conference}

\end{document}